# Intrinsic Grassmann Averages for Online Linear, Robust and Nonlinear Subspace Learning

Rudrasis Chakraborty[1], Søren Hauberg[2] and Baba C. Vemuri[1]

[1]Department of CISE, University of Florida, FL 32611, USA
[2]Technical University of Denmark, Kgs. Lyngby, Denmark.
[1]{rudrasischa, baba.vemuri}@gmail.com   [2]{sohau}@dtu.dk *

## Abstract

Principal Component Analysis (PCA) and Kernel Principal Component Analysis (KPCA) are fundamental methods in machine learning for dimensionality reduction. The former is a technique for finding this approximation in finite dimensions and the latter is often in an infinite dimensional Reproducing Kernel Hilbert-space (RKHS). In this paper, we present a geometric framework for computing the principal linear subspaces in both situations as well as for the robust PCA case, that amounts to computing the intrinsic average on the space of all subspaces: the Grassmann manifold. Points on this manifold are defined as the subspaces spanned by $K$-tuples of observations. The intrinsic Grassmann average of these subspaces are shown to coincide with the principal components of the observations when they are drawn from a Gaussian distribution. We show similar results in the RKHS case and provide an efficient algorithm for computing the projection onto the this average subspace. The result is a method akin to KPCA which is substantially faster. Further, we present a novel online version of the KPCA using our geometric framework. Competitive performance of all our algorithms are demonstrated on a variety of real and synthetic data sets.

*This research was in part funded in part by the NSF grants IIS-1525431 and IIS-1724174 to BCV. SH was supported by a research grant (15334) from VILLUM FONDEN. This project has received funding from the European Research Council (ERC) under the European Union's Horizon 2020 research and innovation programme (grant agreement nº 757360).

# 1 Introduction

Principal component analysis (PCA), a key workhorse of machine learning, can be derived in many ways: Pearson [1] proposed to find the subspace that minimizes the projection error of the observed data; Hotelling [2] instead sought the subspace in which the projected data has maximal variance; and Tipping & Bishop [3] consider a probabilistic formulation where the covariance of normally distributed data is predominantly given by a low-rank matrix. All these derivations lead to the same algorithm. Recently, Hauberg et al. [4, 5] noted that the average of all one-dimensional subspaces spanned by normally distributed data coincides with the leading principal component. They computed the average over the Grassmann manifold of one-dimensional subspaces (cf. Sec. 2). This average was computed very efficiently, but unfortunately their formulation does not generalize to higher-dimensional subspaces.

In this paper, we provide a formulation for estimating the average $K$-dimensional subspace spanned by the observed data, and present a very simple, parameter-free online algorithm for computing this average. When the data is normally distributed, we show that this average subspace coincides with that spanned by the leading $K$ principal components. We further show that our online algorithm has a linear convergence rate. Moreover, since our algorithm is online, it has a linear complexity in terms of the number of samples. Furthermore, we propose an online robust subspace averaging algorithm which can be used to get the leading $K$ robust principal components. Analogous to its non-robust counterpart, it has a linear time complexity in terms of the number of samples. A preliminary conference version of this work was published in [6]. In this article, we generalize this preliminary work to perform online non-linear subspace learning which is an online version of the kernel PCA. In comparison to our preliminary work in [6], this paper contains a more detailed analysis in addition to the generalization akin to kernel PCA [7].

## 1.1 Related Work

In this paper we consider a simple linear dimensionality reduction algorithm that works in an online setting, i.e. only uses each data point once. There are several existing approaches in the literature that tackle the online PCA and the online Robust PCA problems and we discuss some of these approaches here:

*Oja's rule* [8] is a classic online estimator for the leading principal components of a dataset. Given a basis $V_{t-1} \in \mathbf{R}^{D \times K}$ this is updated recursively via $V_t = V_{t-1} + \gamma_t X_t(X_t^T V_{t-1})$ upon receiving the observation $X_t$. Here $\gamma_t$ is the step-size (learning rate) parameter that must be set manually; small values yields slow-but-sure convergence, while larger values may lead to fast-but-unstable convergence.

*EM-PCA* [9] is usually derived for probabilistic PCA, but is easily be adapted to the online setting [10]. Here, the E- and M-steps are given by:

$$\textbf{(E-step)} \ Y_t = (V_{t-1}^T V_{t-1})^{-1}(V_{t-1}^T X_t) \quad (1)$$
$$\textbf{(M-step)} \ \tilde{V}_t = (X_t Y_t^T)(Y_t Y_t^T)^{-1}. \quad (2)$$

The basis is updated recursively via the recursion, $V_t = (1 - \gamma_t)V_{t-1} + \gamma_t \tilde{V}_t$, where $\gamma_t$ is a step-size.

*GROUSE and GRASTA* [11, 12] are online PCA and matrix completion algorithms. GRASTA can be applied to estimate principal subspaces incrementally on subsampled data. Both of these methods are online and use rank-one updation of the principal subspace at each iteration. GRASTA is an online robust subspace tracking algorithm and can be applied to subsampled data and specifically matrix completion problems. He et al. [12] proposed an $\ell_1$-norm based fidelity term that measures the error between the subspace estimate and the outlier corrupted observations. The robustness of GRASTA is attributed to this $\ell_1$-norm based cost. Their formulation of the subspace estimation involves the minimization of a non-convex function in an augmented Lagrangian framework. This optimization is carried out in an alternating fashion using the well known ADMM [13] for estimating a set of parameters involving the weights, the sparse outlier vector and the dual vector in the augmented Lagrangian framework. For fixed estimated values of these parameters, they employ an incremental gradient descent to solve for the low dimensional subspace. Note that the solution obtained is not the optimum of the combined non-convex function of GRASTA.

*Recursive covariance estimation* [14] is straightforward, and the principal components can be extracted via standard eigen-decompositions. Boutsidis et al. [14] consider efficient variants of this idea, and provide elegant performance bounds. The approach does not however scale to high-dimensional data as the covariance cannot practically be stored in memory for situations involving very large data sets as those considered in our work.

Candes et al. [15] formulated Robust PCA (RPCA) as separating a matrix into a low rank ($L$) and a sparse matrix ($S$), i.e., data matrix $X \approx L + S$. They proposed Principal Component Pursuit (PCP) method to robustly find the principal subspace by decomposing into $L$ and $S$. They showed that both $L$ and $S$ can be computed by optimizing an objective function which is a linear combination of nuclear norm on $L$ and $\ell_1$ norm on $S$. Recently, Lois et al. [16] proposed an online RPCA algorithm to solve two interrelated problems, matrix completion and online robust subspace estimation. Candes et al. [15] have some assumptions including a good estimate of the initial subspace and that the basis of the subspace is dense. Though the authors have shown correctness of their algorithm under these assumptions, they are often not practical. In another recent work, Ha and Barber [17] proposed an online RPCA algorithm when $X = (L+S)C$ where $C$ is a data compression matrix. They proposed an algorithm to extract $L$ and $S$ when the data $X$ are compress sensed. This problem is quite interesting in its own right but not something pursued in our work presented here. Feng et al. [18] solved RPCA using a stochastic optimization approach. They show that if each observation is bounded, then their solution converges to the batch mode RPCA solution, i.e., their sequence of robust subspaces converges to the "true" subspace. Hence, they claimed that as the "true subspace" (subspace recovered by RPCA) is robust, so is their online estimate. Though their algorithm is online, the optimization steps ( Algorithm 1 in [18]) are computationally expensive for high-dimensional data. In an earlier paper, Feng et al. [19] proposed a deterministic approach to solve RPCA (dubbed DHR-PCA) for high-dimensional data. They also showed that they can achieve maximal robustness, i.e., a breakdown point of 50%. They proposed a robust computation of the variance matrix and then performed PCA on this matrix to get robust PCs. This algorithm is suitable for very high dimensional data. As most of our real applications in this paper are in very high dimensions, we find DHR-PCA to be well suited to carry out comparisons with. For a further literature study on this rich topic, we refer the reader to [20, 21].

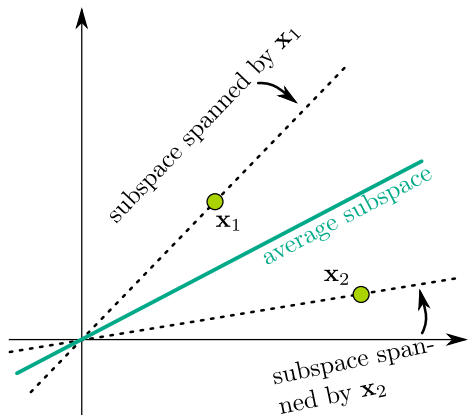

Figure 1: The average of two subspaces.

In this paper, we propose an online subspace averaging algorithm that we also extend to an algorithm akin to the kernel PCA, i.e., to compute non-linear subspaces. We show that with the popular kernel trick, we can extend our subspace averaging algorithm to compute a non-linear average subspace. But, because of the infinite dimensionality of RKHS, it is not computationally feasible to make this an online algorithm. We however resolve this problem by a finite approximation of the kernel using the method proposed in [22] leading to an online non-linear subspace averaging algorithm. The past work in this context includes extension of Oja's rule to perform kernel PCA [7]. Honeine [23] proposed an online kernel PCA algorithm. They pointed out that as the principal vector is a linear combination of the kernel functions associated with the available training data, it becomes a bottleneck in making the kernel PCA online. They overcome this problem by controlling the order of the model. The algorithm starts with a set of preselected kernel functions. Upon the arrival of a new observation, the algorithm decides whether to include or discard the observation from the set of kernel functions. Thus, by restricting the number of kernel functions, they made their kernel PCA algorithm an online algorithm.

*Motivation for our work:* Our work is motivated by the work presented by Hauberg et al. [4], who showed that for a data set drawn from a zero-mean multivariate Gaussian distribution, the average subspace spanned by the data coincides with the leading principal component. This idea is sketched in Fig. 1. Given, $\{\mathbf{x}_i\}_{i=1}^N \subset \mathbf{R}^D$, the 1-dimensional subspace spanned by each $\mathbf{x}_i$ is a point on the Grassmann manifold (Sec. 2). Hauberg et al. then computed the average of these subspaces on the Grassmannian using an "extrinsic" metric, i.e. the Euclidean, distance. Besides the theoretical insight, this formulation gave rise to highly efficient algorithms. Unfortunately, the extrinsic approach is limited to one-dimensional subspaces, and Hauberg et al. resort to deflation methods to estimate higher dimensional subspaces. We overcome this limitation by using an intrinsic metric, extend the theoretical analysis of Hauberg et al., and provide an efficient online algorithm for subspace estimation. We further propose an online non-linear and robust subspace averaging algorithm akin to online KPCA and RPCA respectively. We also present a proof that in the limit, our proposed online robust intrinsic averaging method returns the first $K$ robust principal components.

# 2 An Online Linear Subspace Learning Algorithm

In this section, we present an efficient online linear subspace learning algorithm for finding the principal components of a data set. We first briefly discuss the geometry of the Riemannian manifold of $K$-dimensional linear subspaces in $\mathbf{R}^D$. Then, we will present an online algorithm using the geometry of these subspaces to get the first $K$ principal components of the $D$-dimensional data vectors.

## 2.1 The Geometry of Subspaces

The Grassmann manifold (or the Grassmannian) is defined as the set of all $K$-dimensional linear subspaces in $\mathbf{R}^D$ and is denoted by $\mathsf{Gr}(K, D)$, where $K \in \mathbf{Z}^+$, $D \in \mathbf{Z}^+$, $D \geq K$. A special case of the Grassmannian is when $K = 1$, i.e., the space of one-dimensional subspaces of $\mathbf{R}^D$, which is known as the *real projective space* (denoted by $\mathbf{R}P^D$). A point $\mathcal{X} \in \mathsf{Gr}(K, D)$ can be specified by a basis, $X$, i.e., a set of $K$ linearly independent vectors in $\mathbf{R}^D$ (the columns of $X$) that spans $\mathcal{X}$. We have $\mathcal{X} = \mathsf{Col}(X)$ if $X$ is a basis of $\mathcal{X}$, where $\mathsf{Col}(.)$ is the column span operator. The set of all $D \times K$ matrices of rank $K$, $K < D$ is defined as the Stiefel manifold $\mathsf{St}(K, D)$. Let $T\mathsf{St}(K, D) = \bigcup_X T_X \mathsf{St}(K, D)$ be the tangent bundle on $\mathsf{St}(K, D)$. Now, consider a Riemannian metric $\tilde{g} : T\mathsf{St}(K, D) \times T\mathsf{St}(K, D) \to \mathbf{R}$ on $\mathsf{St}(K, D)$ defined as follows: Given $\tilde{U}, \tilde{V} \in T\mathsf{St}(K, D)$, $\tilde{g}_X(\tilde{U}_X, \tilde{V}_X) = \mathrm{trace}((X^T X)^{-1} \tilde{U}_X^T \tilde{V}_X)$. It is easy to see that the general linear Group $\mathsf{GL}(k)$ acts isometrically, freely and properly on $\mathsf{St}(K, D)$. Moreover, $\mathsf{Gr}(K, D)$ can be identified with the quotient space $\mathsf{St}(K, D)/\mathsf{GL}(K)$. Hence, the projection map $\Pi : \mathsf{St}(K, D) \to \mathsf{Gr}(K, D)$ is a *Riemannian submersion*, where $\Pi(X) := \mathsf{Col}(X)$. Moreover, the triplet $(\mathsf{St}(K, D), \Pi, \mathsf{Gr}(K, D))$ is a principal fiber bundle.

At each point $X \in \mathsf{St}(K, D)$, we can define *vertical space*, $\mathcal{V}_X \subset T_X \mathsf{St}(K, D)$ to be $\mathrm{Ker}(\Pi_{*X})$. Further, given $\tilde{g}$, we define the *horizontal space*, $\mathcal{H}_X$ to be the $\tilde{g}$-orthogonal complement of $\mathcal{V}_X$. Now, from the theory of principal bundles, for every vector field $U$ on $\mathsf{Gr}(K, D)$, we define the *horizontal lift* of $U$ to be the unique vector field $\tilde{U}$ on $\mathsf{St}(K, D)$ for which $\tilde{U}_X \in \mathcal{H}_X$ and $\Pi_{*X} \tilde{U}_X = U_{\Pi(X)}$, $\forall X \in \mathsf{St}(K, D)$. As, $\Pi$ is a Riemannian submersion, the isomorphism $\Pi_{*X}|_{\mathcal{H}_X} : \mathcal{H}_X \to T_{\Pi(X)} \mathsf{Gr}(K, D)$ is an isometry from $(\mathcal{H}_X, \tilde{g}_X)$ to $(T_{\Pi(X)} \mathsf{Gr}(K, D), g_{\Pi(X)})$, where $g$ is the Riemannian metric on $\mathsf{Gr}(K, D)$ defined as:

$$g_{\Pi(X)}(U_{\Pi(X)}, V_{\Pi(X)}) = \tilde{g}_X(\tilde{U}_X, \tilde{V}_X)$$
$$= \mathrm{trace}((X^T X)^{-1} \tilde{U}_X^T \tilde{V}_X) \quad (3)$$

where, $U, V \in T_{\Pi(X)} \mathsf{Gr}(k, n)$ and $\Pi_{*X} \tilde{U}_X = U_{\Pi(X)}$, $\Pi_{*X} \tilde{V}_X = V_{\Pi(X)}$, $\tilde{U}_X \in \mathcal{H}_X$ and $\tilde{V}_X \in \mathcal{H}_X$. Given $\mathcal{X}, \mathcal{Y} \in \mathsf{Gr}(K, D)$, with their respective orthonormal basis $X$ and $Y$, the unique geodesic $\Gamma_\mathcal{X}^\mathcal{Y} : [0, 1] \to \mathsf{Gr}(K, D)$ between $\mathcal{X}$ and $\mathcal{Y}$ is given by:

$$\Gamma_\mathcal{X}^\mathcal{Y}(t) = \mathrm{span}\left(X\hat{V}\cos(\Theta t) + \hat{U}\sin(\Theta t)\right) \quad (4)$$

with $\Gamma_\mathcal{X}^\mathcal{Y}(0) = \mathcal{X}$ and $\Gamma_\mathcal{X}^\mathcal{Y}(1) = \mathcal{Y}$, where, $\hat{U}\hat{\Sigma}\hat{V}^T = (I - XX^T)Y(X^T Y)^{-1}$ is the "thin" Singular value decomposition (SVD), (i.e., $\hat{U}$ is $D \times K$ and $\hat{V}$ is $K \times K$ column orthonormal matrix, and $\hat{\Sigma}$ is $K \times K$ diagonal matrix), and $\Theta = \arctan \hat{\Sigma}$. The length of the geodesic constitutes the geodesic distance on $\mathsf{Gr}(K, D)$, $d : \mathsf{Gr}(K, D) \times \mathsf{Gr}(K, D) \to \mathbf{R}^+ \cup \{0\}$ which is as follows: Given $\mathcal{X}, \mathcal{Y}$ with respective orthonormal bases $X$ and $Y$,

$$d(\mathcal{X}, \mathcal{Y}) \triangleq \sqrt{\sum_{i=1}^{K} (\sigma_i)^2}, \quad (5)$$

where $\bar{U} \Sigma \bar{V}^T = X^T Y$ be the SVD of $X^T Y$, and, $[\sigma_1, \ldots, \sigma_K] = \mathrm{diag}(\Sigma)$. Here $\sigma_i$ is known as the $i^{th}$ *principal angle* between subspace $\mathcal{X}$ and $\mathcal{Y}$.

## 2.2 The Intrinsic Grassmann Average (IGA)

We now consider *intrinsic averages*[1] (IGA) on the Grassmannian. To examine the existence and uniqueness of IGA, we need to define an open ball on the Grassmannian.

---

[1] These are also known as Fréchet means [24, 25].

**Definition 1** (Open ball). *An open ball $\mathcal{B}(\mathcal{X}, r) \subset \mathsf{Gr}(K, D)$ of radius $r > 0$ centered at $\mathcal{X} \in \mathsf{Gr}(K, D)$ is defined as*

$$\mathcal{B}(\mathcal{X}, r) = \{\mathcal{Y} \in \mathsf{Gr}(K, D) | d(\mathcal{X}, \mathcal{Y}) < r\}. \quad (6)$$

Let $\kappa$ be the supremum of the sectional curvature in the ball. Then, we call this ball "regular" [26] if $2r\sqrt{\kappa} < \pi$. Using the results in [27], we know that, for $\mathbf{R}P^D$ with $D \geq 2$, $\kappa = 1$, while for general $\mathsf{Gr}(K, D)$ with $\min(K, D) \geq 2$, $0 \leq \kappa \leq 2$. So, on $\mathsf{Gr}(K, D)$ the radius of a "regular geodesic ball" is $< \pi/2\sqrt{2}$, for $\min(K, D) \geq 2$ and on $\mathbf{R}P^D$, $D \geq 2$, the radius is $< \pi/2$.

Let $\mathcal{X}_1, \ldots, \mathcal{X}_N$ be independent samples on $\mathsf{Gr}(K, D)$ drawn from a distribution $P(\mathcal{X})$, then we can define an *intrinsic average* (FM), [24, 25], $\mathcal{M}^*$ as:

$$\mathcal{M}^* = \operatorname*{argmin}_{\mathcal{M} \in \mathsf{Gr}(K, D)} \sum_{i=1}^{N} d^2(\mathcal{M}, \mathcal{X}_i). \quad (7)$$

As mentioned before, on $\mathsf{Gr}(K, D)$, IGA exists and is unique if the support of $P(\mathcal{X})$ is within a "regular geodesic ball" of radius $< \pi/2\sqrt{2}$. Note that for $\mathbf{R}P^D$, we can choose this bound to be $\pi/2$. In the rest of the paper, we have assumed that data points on $\mathsf{Gr}(K, D)$ are within a "regular geodesic ball" of radius $< \pi/2\sqrt{2}$ unless otherwise specified. With this assumption, the IGA is unique. *Note that this assumption is only needed for proving the theorem presented below.*

The IGA may be computed using a Riemannian steepest descent, but this is computationally expensive and requires selecting a suitable step-size [28]. Recently Chakraborty et al. [29] proposed a simple and efficient recursive/inductive Fréchet mean estimator given by:

$$\mathcal{M}_1 = \mathcal{X}_1$$
$$\mathcal{M}_{k+1} = \Gamma_{\mathcal{M}_k}^{\mathcal{X}_{k+1}}\left(\frac{1}{k+1}\right), \quad \forall k \geq 1 \quad (8)$$

This approach only needs a single pass over the data set to estimate the IGA. Consequently, Eq. 8 has linear complexity in the number of observations. Furthermore, it is truly a online algorithm since each iteration only needs one new observation.

Equation 8 merely performs repeated geodesic interpolation, which is analogous to standard recursive estimators of Euclidean averages: Consider observations $\mathbf{x}_k \in \mathbf{R}^D, k = 1, \ldots, N$. Then the Euclidean average can be computed recursively by moving an appropriate distance away from the $k^{th}$ estimator $\mathbf{m}_k$ towards $\mathbf{x}_{k+1}$ on the straight line joining $\mathbf{x}_{k+1}$ and $\mathbf{m}_k$. The inductive algorithm (8) for computing the IGA works in the same way and is entirely based on traversing geodesics in $\mathsf{Gr}(K, D)$ and without requiring any optimization.

**Theorem 1.** *(Weak Consistency [29]) Let $\mathcal{X}_1, \ldots, \mathcal{X}_N$ be samples on $\mathsf{Gr}(K, D)$ drawn from a distribution $P(\mathcal{X})$. Then, $\mathcal{M}_N$ (8) converges to the IGA of $\{\mathcal{X}_i\}_{i=1}^N$ in probability as $N \to \infty$.*

**Theorem 2.** *(Convergence rate) Let $\mathcal{X}_1, \ldots, \mathcal{X}_N$ be samples on $\mathsf{Gr}(K, D)$ drawn from a distribution $P(\mathcal{X})$. Then Eq. 8 has a linear convergence rate.*

*Proof.* Let $\mathcal{X}_1, \ldots, \mathcal{X}_N$ be the samples drawn from a distribution $P(\mathcal{X})$ on $\mathsf{Gr}(K, D)$. Let $\mathcal{M}$ be the IGA of $\{\mathcal{X}_i\}$. Then, using the triangle inequality we have,

$$\begin{aligned}
d(\mathcal{M}_k, \mathcal{M}) &\leq d(\mathcal{M}_{k-1}, \mathcal{M}_k) + d(\mathcal{M}_{k-1}, \mathcal{M}) \\
&= \frac{1}{k} d(\mathcal{M}_{k-1}, \mathcal{X}_k) + d(\mathcal{M}_{k-1}, \mathcal{M}) \\
&\leq \frac{1}{k}\Big(d(\mathcal{M}_{k-1}, \mathcal{M}) + d(\mathcal{X}_k, \mathcal{M})\Big) + \\
&\quad d(\mathcal{M}_{k-1}, \mathcal{M})
\end{aligned}$$

Hence,

$$\frac{d(\mathcal{M}_k, \mathcal{M})}{d(\mathcal{M}_{k-1}, \mathcal{M})} \leq \left(1 + \frac{1}{k} + \frac{d(\mathcal{X}_k, \mathcal{M})}{k \, d(\mathcal{M}_{k-1}, \mathcal{M})}\right)$$

Since $d(\mathcal{M}_{k-1}, \mathcal{M})$ and $d(\mathcal{X}_k, \mathcal{M})$ are finite as $\{\mathcal{X}_i\}$s are within a geodesic ball of finite radius, with $k \to \infty$, $\frac{d(\mathcal{M}_k, \mathcal{M})}{d(\mathcal{M}_{k-1}, \mathcal{M})} \leq 1$. But, the equality holds only if $\mathcal{M}$ lies on the geodesic between $\mathcal{M}_{k-1}$ and $\mathcal{X}_k$. Let, $l < \infty$ and let $\mathcal{M}$ lies on the geodesic between $\mathcal{M}_{k-1}$ and $\mathcal{X}_k$ for some $k > l$. As $\mathcal{M}$ is fixed, using induction, one can easily show that $\mathcal{M}$ can not lie on the same geodesic for all $k > l$. Hence, $r < 1$ and the convergence rate is linear. ■

## 2.3 Principal Components as Grassmann Averages

Following Hauberg et al. [4] we phrase linear dimensionality reduction as an averaging problem on the Grassmannian. We consider an intrinsic Grassmann average (IGA), which allows us to reckon with $K > 1$ dimensional subspaces. We then propose an *online linear subspace learning* and show that for the zero-mean Gaussian data, the expected IGA on $\mathsf{Gr}(K, D)$, i.e., expected $K$-dimensional linear subspace, coincides with the first $K$ principal components.

Given $\{\mathbf{x}_i\}_{i=1}^N$, the algorithm to compute the IGA that produces the leading $K$-dimensional principal subspace is sketched in Algorithm 1.

---
**Algorithm 1:** The IGA algorithm to compute PCs

**Input**: $\{\mathbf{x}_i\}_{i=1}^N \subset \mathbf{R}^D$, $K > 0$
**Output**: $\{\mathbf{v}_1, \ldots, \mathbf{v}_K\} \subset \mathbf{R}^D$
1 Partition the data $\{\mathbf{x}_j\}_{j=1}^N$ into blocks of size $D \times K$ ;
2 Let the $i^{th}$ block be denoted by, $X_i = [\mathbf{x}_{i1}, \ldots, \mathbf{x}_{iK}]$ ;
3 Orthogonalize each block and let the orthogonalized block be denoted by $X_i$ ;
4 Let the subspace spanned by each $X_i$ be denoted by $\mathcal{X}_i \in \mathsf{Gr}(K, D)$ ;
5 Compute IGA, $\mathcal{M}^*$, of $\{\mathcal{X}_i\}$ ;
6 Return the $K$ columns of an orthogonal basis of $\mathcal{M}^*$; these span the principal $K$-subspace.

---

Let $\{\mathcal{X}_i\}$ be the set of $K$-dimensional subspaces as constructed by IGA in Algorithm 1. Moreover, assume that the maximum principal angle between $\mathcal{X}_i$ and $\mathcal{X}_j$ is $< \pi/2\sqrt{2}$, $\forall i \neq j$. The above condition is assumed to ensure that the IGA exists and is unique on $\mathsf{Gr}(K, D)$. This condition can be ensured if the angle between $\mathbf{x}_l$ and $\mathbf{x}_k$ is $< \pi/2\sqrt{2}, \forall \mathbf{x}_l, \mathbf{x}_k$ belong to different blocks. For $\mathbf{x}_l, \mathbf{x}_k$ in a same block, the angle must be $< \pi/2$. *Note that, this assumption is needed to prove Theorem 3. In practice, even if IGA is not unique, we find a local minimizer of Eq. 7 [24], which serves as the principal subspace.*

**Theorem 3.** *(Relation between IGA and PCA)*
*Let us assume that $\mathbf{x}_i \sim \mathcal{N}(\mathbf{0}, \Sigma)$, $\forall i$. Using the same notations as above, the $j^{th}$ column of $M$ is the $j^{th}$ principal vector of $\{\mathbf{x}_i\}_{i=1}^N$, $j = 1, \ldots, K$, i.e., $M$ spans the principal $K$-subspace, $\mathcal{M}^*$, where $\mathcal{M}^*$ is defined as in Eq. 7.*

*Proof.* Let $X_i$ be the corresponding orthonormal basis of $\mathcal{X}_i$, i.e., $X_i$ spans $\mathcal{X}_i$, for all $i$. The IGA, $\mathcal{M}^*$ can be computed using Eq. 7. Let, $X_i = [\mathbf{x}_{i1} \ldots \mathbf{x}_{iK}]$ where $\mathbf{x}_{ij}$ be samples drawn from $N(\mathbf{0}, \Sigma)$ (Gaussian distribution with $\mathbf{0}$ mean and covariance $\Sigma$). Let, $M = [M_1 \ldots M_K]$ be an orthonormal basis of $\mathcal{M}^*$. The distance between $\mathcal{X}_i$ and $\mathcal{M}^*$ is defined as $d^2(\mathcal{X}_i, \mathcal{M}^*) = \sum_{j=1}^K (\arccos((S_i)_{jj}))^2$, where $\bar{U}_i S_i \bar{V}_i^T = M^T X_i$ be the SVD, and $(S_i)_{jj} \geq 0$ (we use $(A)_{lm}$ to denote $(l, m)^{th}$ entry of matrix $A$). As arccos is a decreasing function and a bijection on $[0, 1]$, we can write an alternative form of Eq. 7 as follows:

$$\mathcal{M}^* = \underset{\mathcal{M}}{\arg\max} \sum_{i=1}^N \sum_{j=1}^K ((S_i)_{jj})^2 \qquad (9)$$

In fact the above alternative form can also be derived using a Taylor's expansion of RHS of Eq. 7. Note that, in the above equation $S_i$ is a function of $M$. It is easy to see that $(M^T X_i)_{lm} \sim \mathcal{N}(0, \sigma^2_{M_l})$, $l = 1, \ldots, K$, $m = 1, \ldots, K$. Also, $(M^T X_i \bar{V}_i)_{lm} \sim \mathcal{N}(0, \sigma^2_{M_l})$, $l = 1, \ldots, K, m = 1, \ldots, K$ as $\bar{V}_i$ is orthogonal. Thus, $(S_i)_{ll} = (\bar{U}_i^T M^T X_i \bar{V}_i)_{ll} \sim \mathcal{N}(0, \sigma^2_{\bar{U}_{il} M_l})$. So, $\sum_{j=1}^K (S_i)_{jj}^2 \sim \Gamma(\frac{1}{2} \sum_{j=1}^K \sigma^2_{\bar{U}_{ij} M_{ij}}, 2)$ and $E[\sum_{j=1}^K (S_i)_{jj}^2] = \sum_{j=1}^K \sigma^2_{\bar{U}_{ij} M_j}$. To compute the expected IGA, we take the expectation of Eq. 9. Now, in order to maximize $E[\sum_{j=1}^K (S_i)_{jj}^2] = \sum_{j=1}^K \sigma^2_{\bar{U}_{ij} M_j}$, $\bar{U}_{ij}$ should be the left singular vectors of $M^T X_i$, and $M_j$ should be the $j^{th}$ eigenvector of $\Sigma$, $\forall j = 1, \ldots, K$. Hence, $M$ spans the principal subspace, $\mathcal{M}^*$. ∎

Now, using Theorem 1 and Theorem 3, we replace the line 5 of the IGA Algorithm 1 by Eq. 8 to get an *online subspace learning* algorithm that we call, *Recursive IGA (RIGA)*, to compute leading $K$ principal components, $K \geq 1$.

The key advantages of our proposed RIGA algorithm to compute PCs are as follows:

1. In contrast to work in [4], "IGA" will return the first $K$ PCs, $K \geq 1$.

2. Using the recursive computation of the FM in IGA leads to RIGA, an online PC computation algorithm. Moreover, Theorem 1 ensures the convergence of "RIGA" to "IGA". Hence, our proposed RIGA is an online PCA algorithm.

3. Unlike previous online PCA algorithms, RIGA is parameter free.

4. By virtue of Theorem 2, RIGA converges at a linear rate.

## 3 A Kernel Extension

In this section, we extend RIGA to perform the principal component analysis in a Reproducing Kernel Hilbert Space (RKHS). We extend RIGA to obtain an efficient nonlinear subspace estimator in RKHS akin to kernel PCA [7] and dub our algorithm *Kernel RIGA (KRIGA)*. The key issue to be addressed in KRIGA is that, in order to perform IGA in the RKHS, we will need to cope with an infinite-dimensional Grassmannian. Fortunately, we observe that the distance between two subspaces in RKHS is same as the distances between span of the coefficient matrices with respect to an orthogonal basis. Hence, instead of performing IGA on the subspaces in RKHS, we will perform IGA of the span of the coefficients, which are finite dimensional. The IGA is then computable using the kernel trick. *A key advantage of KRIGA is that it does not require an eigen-decomposition of the Gram matrix.* Furthermore, we extend this formulation to propose an *online* KRIGA algorithm by approximating the kernel function.

### 3.1 Deriving the Kernel Recursive Intrinsic Grassmann Average (KRIGA)

Let $X = \{\mathbf{x}_1, \ldots, \mathbf{x}_N\}$, where $\mathbf{x}_i \in \mathbf{R}^D$, for all $i$. We seek $K$ principal components, $K \leq D$. Let $\mathcal{K}(.,.)$ be the kernel associated with RKHS $H$ and let $\phi(X) = [\phi(\mathbf{x}_1), \ldots, \phi(\mathbf{x}_N)]$, where $\phi : X \to H$. Let $\widetilde{\phi}(X) = [\widetilde{\phi}(\mathbf{x}_1), \ldots, \widetilde{\phi}(\mathbf{x}_N)]$ be the orthogonalization of $\phi(X)$. Since each $\phi(\mathbf{x}_i) \in \text{Col}(\widetilde{\phi}(X))$ we have $\phi(\mathbf{x}_i) = \widetilde{\phi}(X)\langle\widetilde{\phi}(X), \phi(\mathbf{x}_i)\rangle_H$, where,

$$\langle\widetilde{\phi}(X), \phi(\mathbf{x}_i)\rangle_H = [\langle\widetilde{\phi}(\mathbf{x}_1), \phi(\mathbf{x}_i)\rangle_H, \ldots, \langle\widetilde{\phi}(\mathbf{x}_N), \phi(\mathbf{x}_i)\rangle_H]^t. \quad (10)$$

Here, $\langle.,.\rangle_H$ is the inner product in the RKHS.

Let $Y_l = [\phi(\mathbf{x}_{K(l-1)+1}), \ldots, \phi(\mathbf{x}_{Kl})] = \widetilde{\phi}(X)\langle\widetilde{\phi}(X), Y_l\rangle_H$, where $C_l = \langle\widetilde{\phi}(X), Y_l\rangle_H$. Here, $(C_l)_{ij} = \langle\widetilde{\phi}(\mathbf{x}_i), \phi(\mathbf{x}_{K(l-1)+j})\rangle_H$. Note that, $C_l$ is a matrix of dimension $N \times K$.

We observe that $d(\text{Col}(Y_i), \text{Col}(Y_j)) = d(\text{Col}(C_i), \text{Col}(C_j))$ as proved in the following lemma.

**Lemma 1.** *Using the above notations, $d(Col(Y_i), Col(Y_j)) = d(Col(C_i), Col(C_j))$.*

*Proof.* $Y_i = \widetilde{\phi}(X) C_i$ and $Y_j = \widetilde{\phi}(X) C_j$. Now, using (5), we can see that, $d(\text{Col}(Y_i), \text{Col}(Y_j)) = d(\text{Col}(C_i), \text{Col}(C_j))$ *iff* the SVD of $Y_i^T Y_j$ is same as that of $C_i^T C_j$. Now, as $\widetilde{\phi}(X)$ is column orthogonal, hence the result follows. ∎

So, instead of IGA on $\{\mathcal{Y}_l\}$, we can perform IGA on $\{\mathcal{C}_l\}$, where $\mathcal{Y}_l = \text{Col}(Y_l)$ and $\mathcal{C}_l = \text{Col}(C_l)$, $\forall l$.

The orthogonalization $\widetilde{\phi}(X)$ is achieved using the Gram-Schimdt orthogonalization process as follows:

$$\widehat{\phi}(\mathbf{x}_i) = \phi(\mathbf{x}_i) - \sum_{j=1}^{i-1} \langle\phi(\mathbf{x}_i), \widetilde{\phi}(\mathbf{x}_j)\rangle_H \widetilde{\phi}(\mathbf{x}_j) \quad (11)$$

$$\widetilde{\phi}(\mathbf{x}_i) = \widehat{\phi}(\mathbf{x}_i)/\|\widehat{\phi}(\mathbf{x}_i)\|. \quad (12)$$

The elements of $C_l$, i.e., $\langle\widetilde{\phi}(\mathbf{x}_i), \phi(\mathbf{x}_{K(l-1)+j})\rangle_H$ can be computed using the kernel $\mathcal{K}(.,.)$ as given in the Lemma 2 below.

Let the basis matrix of the IGA of $\{\mathcal{C}_l\}$ be $M$. Then the basis matrix of the IGA of $\{\mathcal{Y}_l\}$ is denoted by $\widetilde{U} = \widetilde{\phi}(X)M$. The Columns of $\widetilde{U}$ will give the PCs in RKHS. Note that this tallies with the *Representer theorem* [30], which tells us that the PC in RKHS is a linear combination of the mapped data vectors, $\phi(X)$.

The following corollary holds by virtue of of Theorem 3.

**Corollary 1.** *The expected IGA of $\{\mathcal{Y}_l\}$ is the same as the PC of $X = \{\mathbf{x}_1, \ldots, \mathbf{x}_N\}$ in the Hilbert space $H$.*

The projection of $\mathbf{x}_i$ onto $\widetilde{U}$ is given by $\mathbf{Proj}(\mathbf{x}_i) = \langle \phi(\mathbf{x}_i), \widetilde{\phi}(X) \rangle_H M$, where,

$$\langle \phi(\mathbf{x}_i), \widetilde{\phi}(X) \rangle_H = [\langle \phi(\mathbf{x}_i), \widetilde{\phi}(\mathbf{x}_1) \rangle_H, \ldots, \langle \phi(\mathbf{x}_i), \widetilde{\phi}(\mathbf{x}_N) \rangle_H]. \tag{13}$$

The following Lemma, gives the analytic form of $\langle \widetilde{\phi}(\mathbf{x}_m), \phi(\mathbf{x}_i) \rangle_H$.

**Lemma 2.** $\langle \widetilde{\phi}(\mathbf{x}_m), \phi(\mathbf{x}_i) \rangle_H =$

$$\begin{cases} \frac{\mathcal{K}(\mathbf{x}_m, \mathbf{x}_i) - \sum_{j=1}^{m-1} \langle \phi(\mathbf{x}_m), \widetilde{\phi}(\mathbf{x}_j) \rangle_H \langle \widetilde{\phi}(\mathbf{x}_j), \phi(\mathbf{x}_i) \rangle_H}{\sqrt{\mathcal{K}(\mathbf{x}_m, \mathbf{x}_m) - \sum_{j=1}^{m-1} \langle \phi(\mathbf{x}_m), \widetilde{\phi}(\mathbf{x}_j) \rangle_H}}, & i \geq m \\ 0, & \text{otherwise} \end{cases}$$

*Proof.*

$$\widetilde{\phi}(\mathbf{x}_m) = \phi(\mathbf{x}_m) - \sum_{j=1}^{m-1} \langle \phi(\mathbf{x}_m), \widetilde{\phi}(\mathbf{x}_j) \rangle_H \widetilde{\phi}(\mathbf{x}_j).$$

and

$$\widetilde{\phi}(\mathbf{x}_m) = \widetilde{\phi}(\mathbf{x}_m) / \|\widetilde{\phi}(\mathbf{x}_m)\|.$$

where,

$$\|\widetilde{\phi}(\mathbf{x}_m)\| = \sqrt{\mathcal{K}(\mathbf{x}_m, \mathbf{x}_m) - \sum_{j=1}^{m-1} \langle \phi(\mathbf{x}_m), \widetilde{\phi}(\mathbf{x}_j) \rangle_H}.$$

Now, since $\{\widetilde{\phi}(\mathbf{x}_i)\}_{i=1}^N$ serves as a basis to represent each $\phi(\mathbf{x}_i)$, clearly, $\langle \widetilde{\phi}(\mathbf{x}_m), \phi(\mathbf{x}_i) \rangle_H = 0$ when, $i < m$, $m = 1, \cdots, N$.

So, consider $m \in \{1, \cdots, N\}$, $i \geq m$, then,

$$\langle \widetilde{\phi}(\mathbf{x}_m), \phi(\mathbf{x}_i) \rangle_H = \frac{1}{\|\widetilde{\phi}(\mathbf{x}_m)\|} \left( \mathcal{K}(\mathbf{x}_m, \mathbf{x}_i) - \sum_{j=1}^{m-1} \langle \phi(\mathbf{x}_m), \widetilde{\phi}(\mathbf{x}_j) \rangle_H \langle \widetilde{\phi}(\mathbf{x}_j), \phi(\mathbf{x}_i) \rangle_H \right)$$

∎

Note that, thus far we have implicitly assumed that $\sum_i \phi(\mathbf{x}_i) = 0$, i.e., the mapped data in RKHS is centered. For non-centered data, we have to first center the data. Given the non-centered data, $\{\phi(\mathbf{x}_i\}_{i=1}^N$, let the centered data be denoted by $\{\bar{\phi}(\mathbf{x}_i)\}_{i=1}^N$, where $\bar{\phi}(\mathbf{x}_i) = \phi(\mathbf{x}_i) - \frac{1}{N} \sum_{j=1}^N \phi(\mathbf{x}_j)$. Then, using Lemma 2, the coefficient matrix $C_l$ is computed using

$$\langle \widetilde{\phi}(\mathbf{x}_m), \bar{\phi}(\mathbf{x}_i) \rangle_H = \langle \widetilde{\phi}(\mathbf{x}_m), \phi(\mathbf{x}_i) \rangle_H - \frac{1}{N} \sum_{j=1}^N \langle \widetilde{\phi}(\mathbf{x}_m), \phi(\mathbf{x}_j) \rangle_H \tag{14}$$

Thus, the coefficient matrices $\{C_l\}$ can be computed using only the Grammian matrix $\mathcal{K}$ as can be seen from Eq. 14. In terms of computational complexity, KRIGA takes $\mathcal{O}(N^3 - N^2)$ computations while KPCA takes $\mathcal{O}(N^3)$ computations.

Now, observe that the above KRIGA algorithm (analog to KPCA) is not online because of the following reasons: **(i)** centering step of the data; **(ii)** choice of basis, i.e., $\{\phi(\mathbf{x}_i)\}_{i=1}^N$. Consistent with the online KPCA algorithm, we assume data to be centered. Then, in order to make the above algorithm online, we need to find a predefined basis in RKHS. We will use the idea proposed by Rahimi et al. in [22], to approximate the shift-invariant kernel $\mathcal{K}$. They observed that infinite kernel expansions can be well-approximated using randomly drawn features. For shift-invariant $\mathcal{K}$, this relates to Bochner's lemma [31] as stated below.

**Lemma 3.** $\mathcal{K}$ *is positive definite iff* $\mathcal{K}$ *is the Fourier transform of a non-negative measure,* $\mu(\mathbf{w})$.

This in turn implies the existence of a probability density $p(\mathbf{w}) := \mu(\mathbf{w})/C$, where $C$ is the normalizing constant. Hence,

$$\mathcal{K}(\mathbf{x}, \mathbf{y}) = C \int \exp(-j\mathbf{w}^t(\mathbf{x} - \mathbf{y})) p(\mathbf{w}) d\mathbf{w}$$
$$= C E_{\mathbf{w}} [\cos(\mathbf{w}^t(\mathbf{x} - \mathbf{y}))].$$

The above expectation can be approximated using Monte Carlo methods, more specifically, we will draw $M$ i.i.d. $\mathbf{R}^D$ vectors from $p(\mathbf{w})$ and form matrix $W$ of size $M \times D$. Then, we can approximate $\mathcal{K}(\mathbf{x}, \mathbf{y})$ by,

$$\mathcal{K}(\mathbf{x}, \mathbf{y}) = \psi(W\mathbf{x})^t \psi(W\mathbf{y}),$$

where $\psi(W\mathbf{x}) = \sqrt{C/M} (\cos(W\mathbf{x}), \sin(W\mathbf{x}))^t$. Now, depending on the choice of $\mathcal{K}$, $p(\mathbf{w})$ will change. For example, for the Gaussian RBF kernel, i.e.,

$$\mathcal{K}(\mathbf{x}, \mathbf{y}) = \exp\left(-\frac{\|\mathbf{x} - \mathbf{y}\|^2}{2\sigma^2}\right).$$

**w** should be sampled from $N\left(\mathbf{0}, \text{diag}\left(\sigma^2\right)^{-1}\right)$.

Rahimi et al. in [22] provide a bound on the error in the approximation of the kernel. Now, because of this approximation, in our KRIGA algorithm, we can replace $\phi$ by $\psi$. As $\psi$ is finite dimensional, we will choose the canonical basis in $\mathbf{R}^{2M}$, i.e., replacing $\widetilde{\phi}$ in the above derivation by $\{\mathbf{e}_i\}_{i=1}^{2M}$. This gives us an online KRIGA algorithm using the recursive IGA.

# 4 A Robust Online Linear Subspace Learning Algorithm

In this section, we will propose an online robust PCA algorithm using intrinsic Grassmann averages. Let $\{\mathcal{X}_1, \mathcal{X}_2, \cdots, \mathcal{X}_N\} \subset \mathsf{Gr}(K, D), K < D$ be inside a regular geodesic ball of radius $< \pi/2\sqrt{2}$ s.t., the Fréchet Median (FMe) [32] exists and is unique. Let $X_1, X_2, \cdots, X_N$ be the corresponding orthonormal bases, i.e., $X_i$ spans $\mathcal{X}_i$, for all $i$. The FMe can be computed via the following minimization:

$$\mathcal{M}^* = \arg\min_{\mathcal{M}} \sum_{i=1}^{N} d(\mathcal{X}_i, \mathcal{M}) \qquad (15)$$

With a slight abuse of notation, we use the notation $\mathcal{M}^*(M)$ to denote both the FM and the FMe (and their orthonormal basis). The FMe is robust as was shown by Fletcher et al. [32], hence we call our estimator *Robust IGA* (RoIGA). In the following theorem, we will prove that RoIGA leads to the robust PCA in the limit as the number of the data samples goes to infinity. An algorithm to compute RoIGA is obtained by simply replacing Step 5 of Algorithm 1 by computation of RoIGA via minimization of Eq. 15 instead of Eq. 7. This minimization can be achieved using the Riemannian steepest descent, but instead, here we use the stochastic gradient descent of batch size 5 to compute RoIGA. As at each iteration, we need to store only 5 samples, the algorithm is online. The update step for each iteration of the online algorithm to compute RoIGA (we refer to our online RoIGA algorithm as Recursive RoIGA (RRIGA)) is as follows:

$$\mathcal{M}_1 = \mathcal{X}_1$$
$$\mathcal{M}_{k+1} = \mathrm{Exp}_{\mathcal{M}_k}\left(\frac{\mathrm{Exp}^{-1}_{\mathcal{M}_k}(\mathcal{X}_{k+1})}{(k+1)d(\mathcal{M}_k, \mathcal{X}_{k+1})}\right) \qquad (16)$$

where, $k \geq 1$, Exp and $\mathrm{Exp}^{-1}$ are Riemannian exponential and inverse exponential functions as defined below.

**Definition 2** (Exponential map). *Let $\mathcal{X} \in \mathsf{Gr}(K, D)$. Let $\mathcal{B}(\mathbf{0}, r) \subset T_\mathcal{X}\mathsf{Gr}(K, D)$ be an open ball centered at the origin in the tangent space at $\mathcal{X}$, where $r$ is the injectivity radius [24] of $\mathsf{Gr}(K, D)$. Then, the Riemannian Exponential map is a diffeomorphism $Exp_\mathcal{X} : \mathcal{B}(\mathbf{0}, r) \to \mathsf{Gr}(K, D)$.*

**Definition 3** (Inverse Exponential map). *Since, inside $\mathcal{B}(\mathbf{0}, r)$, Exp is a diffeomorphism, hence the inverse Exponential map is defined and is a map $Exp_\mathcal{X}^{-1} : \mathcal{U} \to \mathcal{B}(\mathbf{0}, r)$, where $\mathcal{U} = Exp_\mathcal{X}(\mathcal{B}(\mathbf{0}, r)) := \{Exp_\mathcal{X}(U)|U \in \mathcal{B}(\mathbf{0}, r)\}$.*

We refer the readers to [33] for the consistency proof of the estimator.

**Theorem 4.** *(Robustness of RoIGA) Assuming the above hypotheses and notations, as $N \to \infty$, the columns of $M$ converge to the robust principal vectors of the $\{\mathbf{x}_i\}_{i=1}^N$, where $M$ is the orthonormal basis of $\mathcal{M}^*$ as defined in Eq. 15.*

*Proof.* Let, $X_i = [\mathbf{x}_{i1} \cdots \mathbf{x}_{iK}]$ and $\mathbf{x}_{ij}$ be i.i.d. samples drawn from $N(\mathbf{0}, \Sigma)$. Let, $M = [M_1 \cdots M_K]$ be an orthonormal basis of $\mathcal{M}$. Recall that the distance between $\mathcal{X}_i$ and $\mathcal{M}$ is defined as $d(\mathcal{X}_i, \mathcal{M}) = \sqrt{\sum_{j=1}^{K}(\arccos((S_i)_{jj}))^2}$, where $\bar{U}_i S_i V_i^T = M^T X_i$ be the SVD, and $(S_i)_{jj} \geq 0$. Since arccos is a decreasing function and is a bijection on $[0, 1]$, we can rewrite Eq. 15 alternatively as follows:

$$\mathcal{M}^* = \arg\max_{\mathcal{M}} \sum_{i=1}^{N} \sqrt{\sum_{j=1}^{K}((S_i)_{jj})^2} \qquad (17)$$

In fact the above alternative form can also be derived using a Taylor expansion of the RHS of Eq. 15.

From the proof of Theorem 3, we know that,

$$\sum_{j=1}^{K}((S_i)_{jj})^2 \sim \Gamma\left(\frac{1}{2}\sum_{j=1}^{K}\sigma^2_{\tilde{U}_{ij}M_j}, 2\right).$$

Hence, $\sqrt{\sum_{j=1}^{K}((S_i)_{jj})^2}$ follows $N_g\left(\frac{1}{2}\sum_{j=1}^{K}\sigma^2_{\tilde{U}_{ij}M_j}, \sum_{j=1}^{K}\sigma^2_{\tilde{U}_{ij}M_j}\right)$, where $N_g$ is the Nakagami distribution [34]. Now, as $N \to \infty$, the RHS of Eq. 17 becomes $E\left[\sqrt{\sum_{j=1}^{K}((S_i)_{jj})^2}\right]$. $E\left[\sqrt{\sum_{j=1}^{K}((S_i)_{jj})^2}\right] = \sqrt{2}\Gamma(\sum_{j=1}^{K}\sigma^2_{\tilde{U}_{ij}M_j} + 0.5)/\Gamma(\sum_{j=1}^{K}\sigma^2_{\tilde{U}_{ij}M_j})$, where $\Gamma$ is the well known gamma function. It is easy to see that as $\Gamma$ is an increasing function, $E\left[\sqrt{\sum_{j=1}^{K}((S_i)_{jj})^2}\right]$ is maximized *iff* $\sum_{j=1}^{K}\sigma^2_{\tilde{U}_{ij}M_j}$ is maximized, i.e., when $M$ spans the principal $K$-subspace.

Now, if we contrast with the objective function of RIGA in Eq. 9, there we had to maximize $E\left[\sum_{j=1}^{K}((S_i)_{jj})^2\right] = \sum_{j=1}^{K}\sigma^2_{\tilde{U}_{ij}M_j}$. Thus, $E\left[\sqrt{\sum_{j=1}^{K}((S_i)_{jj})^2}\right] = \rho(\mathfrak{m}) \triangleq \sqrt{2}\Gamma(\mathfrak{m}+0.5)/\Gamma(\mathfrak{m})$, where $\mathfrak{m} = \sum_{j=1}^{K}\sigma^2_{\tilde{U}_{ij}M_j}$. Hence, the *influence function* [35] of $\rho$ is proportional to $\psi(\mathfrak{m}) \triangleq \frac{\partial E[\sqrt{\sum_{j=1}^{K}((S_i)_{jj})^2}]}{\partial \mathfrak{m}}$ and if we can show that $\lim_{\mathfrak{m}\to\infty}\psi(\mathfrak{m}) = 0$, then we can claim that our objective function in Eq. 17 is robust [35].

Now, $\psi(\mathfrak{m}) = \Gamma(\mathfrak{m})\Gamma(\mathfrak{m}+0.5)\frac{\phi(\mathfrak{m}+0.5)-\phi(\mathfrak{m})}{\Gamma(\mathfrak{m})^2}$, where $\phi$ is the polygamma function [36] of order 0. After some simple calculations, we get,

$$\lim_{\mathfrak{m}\to\infty}(\phi(\mathfrak{m}+0.5)-\phi(\mathfrak{m})) = \lim_{\mathfrak{m}\to\infty}\log(1+1/(2\mathfrak{m}))$$
$$+ \lim_{\mathfrak{m}\to\infty}\sum_{k=1}^{\infty}\left(B_k\left(\frac{1}{k\mathfrak{m}^k} - \frac{1}{k(\mathfrak{m}+0.5)^k}\right)\right)$$
$$= \lim_{\mathfrak{m}\to\infty}\log(1+1/(2\mathfrak{m})) + 0 = 0$$

Here, $\{B_k\}$ are the Bernoulli numbers of the second kind [37]. So, $\lim_{\mathfrak{m}\to\infty}\psi(\mathfrak{m}) = 0$. ∎

We would like to point out that the outlier corrupted data can be modeled using a mixture of independent random variables, $Y_1, Y_2$, where $Y_1 \sim N(\mathbf{0}, \Sigma_1)$ (to model non-outlier data samples) and $Y_2 \sim N(\boldsymbol{\mu}, \Sigma_2)$ (to model outliers), i.e., $(\forall i)$, $\mathbf{x}_i = w_1 Y_1 + (1-w_1)Y_2$, $w_1 > 0$ is generally large, so that the probability of drawing outliers is low. Then as the mixture components are independent, $(\forall i)$, $\mathbf{x}_i \sim N((1-w_1)\boldsymbol{\mu}, w_1^2\Sigma_1 + (1-w_1)^2\Sigma_2)$. A basic assumption in any online PCA algorithm is that data is centered. So, in case the data is not centered (similar to the model of $\mathbf{x}_i$), the first step of PCA would be to centralize the data. But then the algorithm cannot be made online, hence our above assumption that $\mathbf{x}_i \sim N(\mathbf{0}, \Sigma)$ is a common assumption in an online scenario. But, in a general case, after centralizing the data as the first step of PCA, the above theorem is valid.

# 5 Experimental Results

We evaluate the performance of the proposed recursive estimators on both real and synthetic data. Our overall findings are that the RIGA estimator is more accurate than other online linear subspace estimators since it is parameter free. The Kernel RIGA (KRIGA) is found to yield results that are almost identical to Kernel PCA (KPCA) but at a significant reduction in run time. Below we consider RIGA and KRIGA separately.

## 5.1 Online Linear Subspace Estimation

**Baselines:** We compare with Oja's rule and and the online version of EM PCA (Sec. 1.1). For Oja's rule we follow common guidelines and consider step-sizes $\gamma_t = \alpha/D\sqrt{t}$ with $\alpha$-values between 0.005 and 0.2. For EM PCA we follow advice from Cappé [10] and use step-sizes $\gamma_t = 1/t^\alpha$ with $\alpha$-values between 0.6 and 0.9 along with Polyak-Ruppert averaging.

**(Synthetic) Gaussian Data:** Theorem 3 state that the RIGA estimates coincide in expectation with the leading principal subspace when the data is drawn from a zero-mean Gaussian distribution. We empirically verify this for an increasing number of observations drawn from randomly generated zero-mean Gaussians. We measure the *expressed variance* which is the ratio of the variance captured by the estimated subspace to the variance captured by the true principal subspace:

$$\text{Expressed Variance} = \sum_{k=1}^{K} \frac{\sum_{n=1}^{N} \mathbf{x}_n^T \mathbf{v}_k^{(\text{est})}}{\sum_{n=1}^{N} \mathbf{x}_n^T \mathbf{v}_k^{(\text{true})}} \in [0, 1]. \tag{18}$$

An expressed variance of 1 implies that the estimated subspace captures as much variance as the principal subspace. The right panel of Fig. 2 shows the mean ($\pm$ one standard deviation) expressed variance of RIGA over 150 trials. It is evident that for the Gaussian data, the RIGA estimator does indeed converge to the true principal subspace.

A key aspect of any online estimator is that it should be stable and converge fast to a good estimate. Here, we compare RIGA to the above-mentioned baselines. Both Oja's rule and EM-PCA require a step-size to be specified, so we consider a larger selection of such step-sizes. The left panel of Fig. 2 shows the expressed variance as a function of number of observations for different estimators and step-sizes. In Fig. 3, we have comparative performance analysis of EM-PCA, GROUSE, Oja's rule and RIGA. EM-PCA was found to be quite stable with respect to the choice of step-size, though it does not seem to converge to a good estimate. Oja's rule, on the other hand, seems to converge to a good estimate, but its practical performance is critically dependent on the step-size (as evident from Fig. 2). GROUSE is seen to oscillate for small data size however, with a large number of samples, it yields a good estimate. On the other hand, RIGA is parameter-free and is observed to have good convergence properties.

In the right panel of Fig. 3, we perform a stability analysis of GROUSE and RIGA. Here, for a fixed value of $N$, we generate a data matrix and perform 200 independent runs on the data matrix and report the mean ($\pm$ one standard deviation) expressed variance. As can be seen from the figure, RIGA is very stable in comparison to GROUSE.

**Human Body Shape:** Online algorithms are generally well-suited for solving large-scale problems as they, by construction, should have linear time-complexity in the number of observations. As an example we consider a large collection of three-dimensional scans of human body shape [38]. This dataset contains $N = 21862$ meshes which each consist of 6890 vertices in $\mathbf{R}^3$. Each mesh is, thus, viewed as a $D = 6890 \times 3 = 20670$ vector. We estimate a $K = 10$ dimensional principal subspace using Oja's rule, EM PCA and RIGA respectively. The average reconstruction error over all meshes are 16.8 mm for Oja's rule, 1.9 mm for EM PCA, and 1.0 mm for RIGA. *Note that both Oja's rule and EM PCA explicitly minimize the reconstruction error, while RIGA does not but yet outperforms the baseline methods.* We speculate that this is due to RIGA's excellent convergence properties and it being a parameter free algorithm is not bogged down by the hard problem of step-size tuning confronted in the baseline algorithms used here.

**Santa Claus Conquers the Martians:** We now consider an even larger scale experiment and consider all frames of the motion picture *Santa Claus Conquers the Martians (1964)*[2]. This consist of $N = 145,550$ RGB frames of size $320 \times 240$, corresponding to an image dimension of $D = 230,400$. We estimate a $K = 10$ dimensional subspace using Oja's rule, EM PCA and RIGA respectively. Again, we measure the accuracy of the different estimators via the reconstruction error. Pixel intensities are scaled to be between 0 and 1. Oja's rule gives an average reconstruction error of 0.054, EM PCA gives 0.025, while RIGA gives 0.023. Here RIGA and EM PCA gives roughly equally good results, with a slight advantage to RIGA. Oja's rule does not fare as well. As with the shape data, it is interesting to note that RIGA outperforms the other baseline methods on the error measure that they optimize even though RIGA optimizes a different measure.

### 5.2 Nonlinear Subspace Estimation

In this section, we analyze comparative performance of our proposed KRIGA and online KRIGA with KPCA as the baseline. In our experiments, we used a Gaussian kernel with $\sigma = 1$. The performance is compared in terms of the time required and *average reconstruction error (ARE)* [7]. In Fig. 4, we present a synthetic experiment, where the data generated is in the form of three concentric circles. We can see that both KPCA and online KRIGA yield similar cluster separation. As expected, we can see that with very few dimensions for approximation (i.e., with small $M$), the performance of online KRIGA is poor. Similar observation can be made from Fig. 5, where with $M = 500$, we get almost as good result as KPCA.

---

[2] https://archive.org/details/SantaClausConquerstheMartians1964

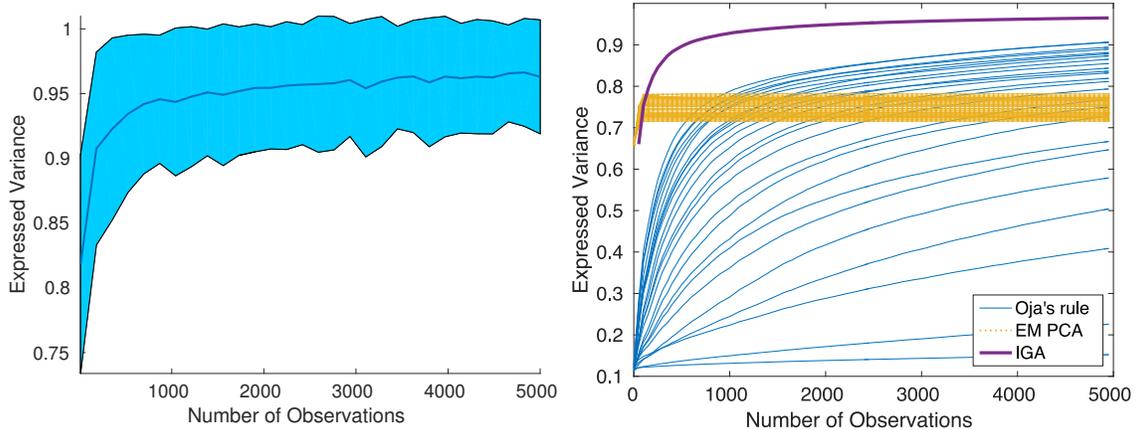

Figure 2: Expressed variance as a function of number of observations. *Left:* The mean and one standard deviation of the RIGA estimator computed over 150 trials. In each trial data are generated in $\mathbf{R}^{50}$ and we estimate a $K = 2$ dimensional subspace. *Right:* The performance of different estimators for varying step-sizes. Here data are generated in $\mathbf{R}^{250}$ and we estimate a $K = 20$ dimensional subspace. In both experiments, we observe similar trends with other values of $D$ and $K$.

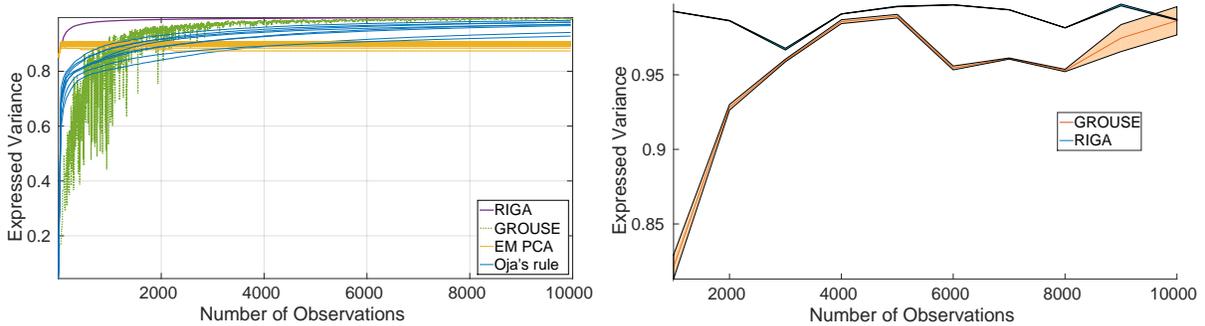

Figure 3: Expressed variance as a function of number of observations. *Left:* The performance of different estimators. Data are generated in $\mathbf{R}^{250}$ and we set $K = 20$. In both experiments, we observe similar trends with other values of $D$ and $K$. *Right:* Stability analysis comparison of GROUSE and RIGA (for a fixed $N$, we randomly generate a data matrix, $X$, from a Gaussian distribution on $\mathbf{R}^{250}$. we estimate $K = 20$ dimensional subspace and report the mean and one standard deviation over 200 runs on $X$.)

Now, we assess the performance of KRIGA and KPCA in terms of ARE and computation time, based on randomly generated synthetic data. Here, we compare our online KRIGA with KPCA. In order to make a fair comparison, we have used the MATLAB 'eigs' function of KPCA which is significantly faster than KPCA. The results for ARE and computation time are shown in Fig. 7. We can see that our online KRIGA is faster than KPCA with 'eigs' without sacrificing much ARE. For this experiment, we chose $K = 5$. Though, the ARE of KPCA is better than that of KRIGA, we can see from Fig. 6 that with increasing number of PCs, performance of KRIGA is similar to KPCA.

Finally, we tested the KPCA and online KRIGA algorithms on the entire movie, *Santa Claus Conquers the Martians (1964)*, and show the time comparison in Fig. 8. We observe an cubical time growth for KPCA while for the online KRIGA, the time is almost a constant. This demonstrates the scalability of our proposed method.

## 5.3 Robust Subspace Estimation

We now present the comparative experimental evaluation of robust extension (RRIGA). Here we use DHR-PCA and GRASTA as baseline methods and measure the performance using the *reconstruction error (RE)*. We used UCSD anomaly detection database [39] and the Extended YaleB database [40] respectively in this evaluation.

**UCSD anomaly detection database:** This data contains images of pedestrian movement on walkways captured by a stationary mounted camera. The crowd density on the walkway varies from sparse to very crowded. The anomaly includes bikers, skaters, carts, people in wheelchair etc. This database is divided in two sets: "Peds1" (people are walking towards the camera) and "Peds2" (people are walking parallel to the camera plane). In "Peds1" there are 36 training and 34 testing videos where each video contains 180 frames of dimension $158 \times 238$ ($D = 37604$). In "Peds2" there are 12 training and 16 testing videos containing varying samples of dimension $240 \times 360$ ($D = 86400$). The test frames do not have anomalous activities. Some sample frames (with and without outliers) are shown in Fig. 9. We first extract $K$ principal components on the training data (including anomalies) and then compute reconstruction error on the test frames (without anomalies) using the computed principal components. It is expected that if the PC computation technique is robust, the reconstruction error will be good as PCs should not be affected by the anomalies in training samples. In Fig. 10, we compare performance of RRIGA with GRASTA and DHR-PCA in terms of RE and time required by varying $K$ from 1 to 100. In terms of time it is evident that RRIGA is very fast compared to both GRASTA and DHR-PCA. RRIGA also outperforms both DHR-PCA and GRASTA in terms of RE. Moreover, it is evident that RRIGA scales very well both in terms of RE and computation time unlike it's competitors.

**Yale ExtendedB database:** This data contains 2414 face images of 38 subjects. We crop each image to make a $32 \times 32$ images ($D = 1024$). Due to varying lighting condition, some of the face images are shaded/dark and appeared as outliers (this experimental setup is similar to the one in [41]). In Fig. 11 some sample face images (outlier and non-outlier) are shown. One can see that due to poor lighting condition, though the middle face in top row is a face image, it looks completely dark and an outlier. For testing, we have used 142 non-outlier face images of 38 subjects and the rest we used to extract PCs. We report RE (with varying $K$) and time required for both RRIGA, GRASTA and DHR-PCA in Fig. 12. From the figure it is evident that for small number of PCs (i.e., small $K$) RRIGA performs similar to DHR-PCA, while for larger $K$ values, RRIGA outperforms DHR-PCA and GRASTA. In terms of time required, RRIGA is faster than both DHR-PCA and GRASTA.

**Background separation:** Here, we use the popular Wallflower Images dataset [42] to perform background separation from images. In this setup, we use robust PCA on images with varying percentage of noise. And then use the extracted PCs to separate background from a test image. We perform this task on two different sets of data namely the 'Camoflouge' and 'Waving Tree' respectively. In order to perform the training, we chose images with 5% and 10% outliers respectively. Examples of training images, each of dimension 57600, from both the data sets are depicted in Fig. 13.

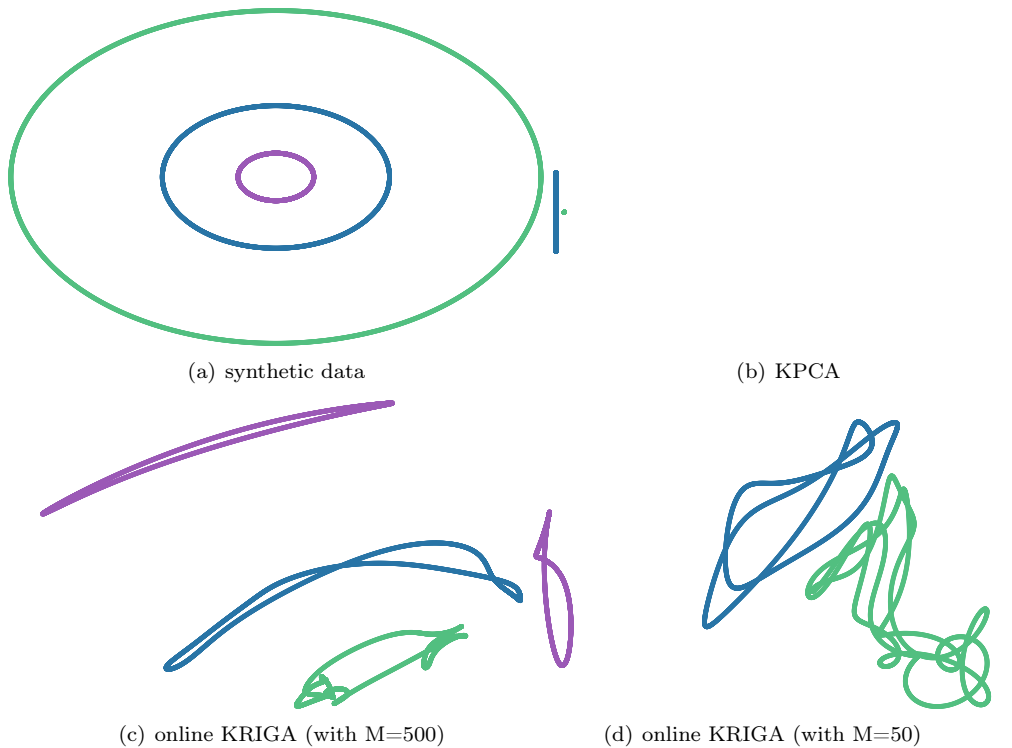

Figure 4: results on synthetic data

We presented the performance of RRIGA, DHR-PCA and standard PCA in Fig. 14. As expected, we can see that the performance of PCA is poor, while RRIGA and DHR-PCA both perform equally good in separating the background for both 5% and 10% outliers in the training. Even when a majority of the image is obstructed by an object, both DHR-PCA and RRIGA performs well. On the 'Camoflouge' data, the performance of DHR-PCA is slightly worse than RRIGA as evident in the form of a black shadow resembling the object present in the background.

# 6 Conclusions

In this paper, we presented a geometric framework to compute principal linear subspaces in finite and infinite dimensional reproducing kernel Hilbert spaces (RKHS). We computed an intrinsic Grassmann average as a proxy for the principal linear subspace and showed that if the samples are drawn from a Gaussian distribution, the intrinsic Grassmann average coincides with the principal subspace in expectation. We further showed that the approach extends to the RKHS setting. A robust version of the online PCA is also presented along with several experiments demonstrating its performance in comparison to the state-of-the-art. The approach has several advantages.

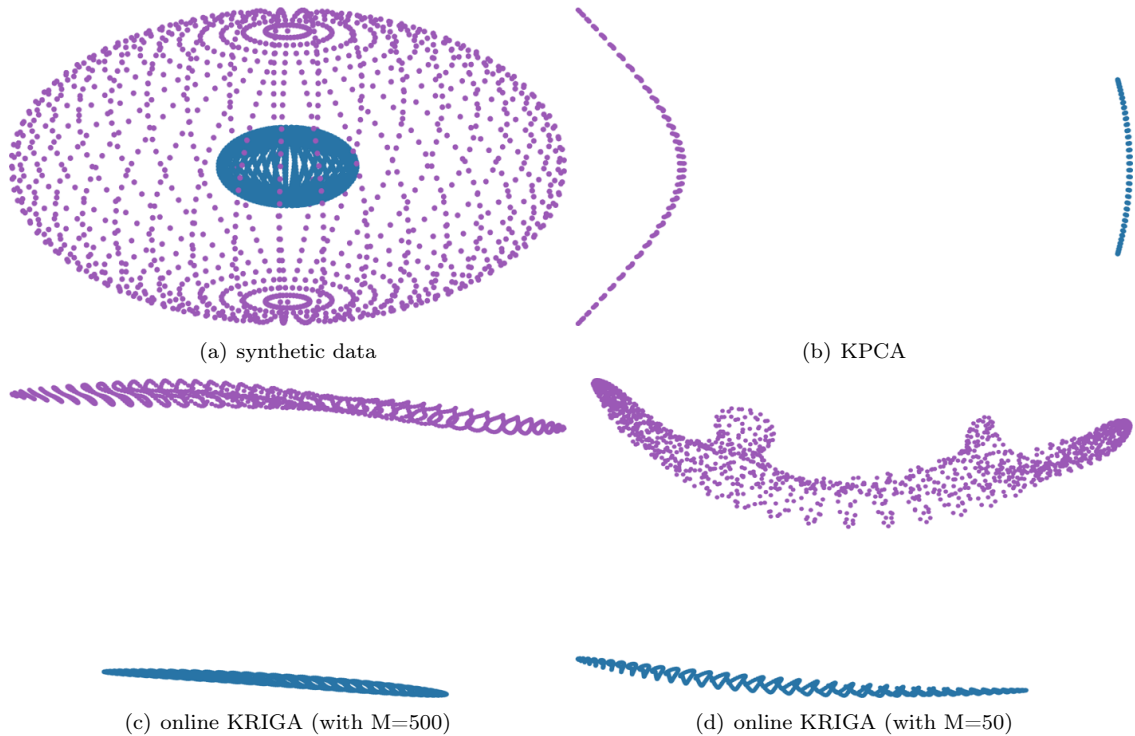

Figure 5: results on 3D synthetic data

Unlike the work by Hauberg et al. in [4], our estimator returns the first $K \geq 1$ components. The proposed algorithm is inherently online, which also makes it scalable to large datasets. We have demonstrated this by performing principal component analysis of an entire Hollywood movie. Theoretically, we proved that the convergence rate of this online algorithm is linear in the number of data points, which indicates that the online estimator converges quickly. Unlike most other online algorithms there are no step-sizes or other parameters to tune; a very useful property in practical settings. We extended the approach to RKHS and thereby provided an algorithm that serves the same purpose as kernel PCA. A benefit of our formulation is that, unlike KPCA, our estimator does not require an eigen-decomposition of the Gram matrix. Empirically, we observed that our algorithm is significantly faster than KPCA while giving similar results.

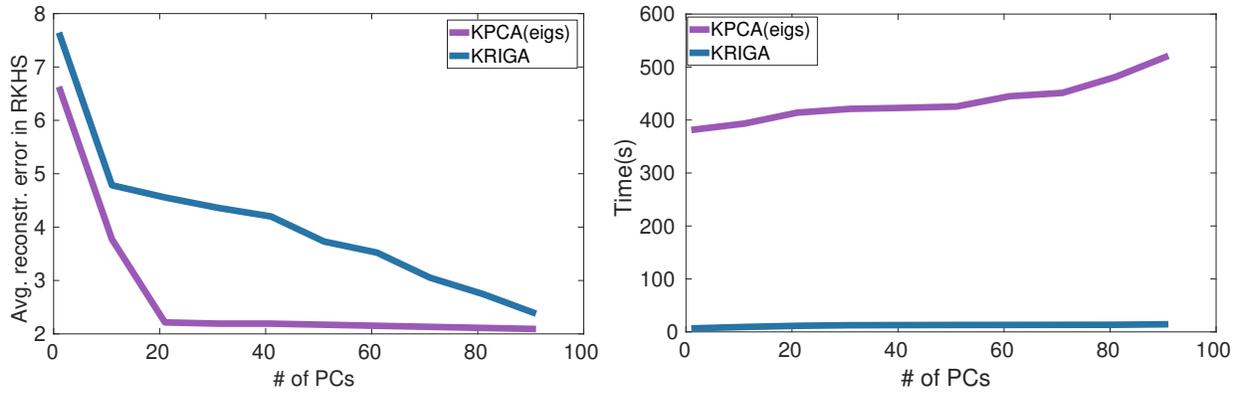

Figure 6: Comparison between KPCA (with eigs) and online KRIGA with varying number of PCs in terms of *Left: Average reconstruction error (ARE)*; *Right:* running time.

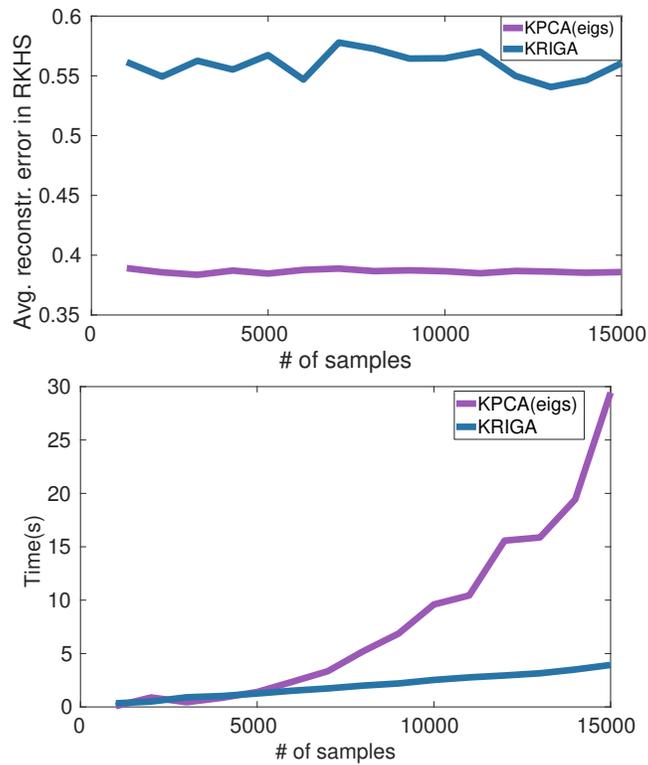

Figure 7: Comparison between KPCA (with eigs) and online KRIGA in terms of *Left: Average reconstruction error (ARE)*; *Right:* running time.

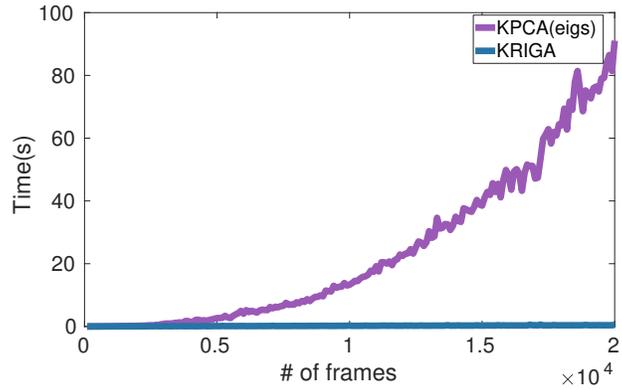

Figure 8: Comparison between KPCA (with eigs) and online KRIGA in terms of running time on an entire movie sampled in 10 FPS.

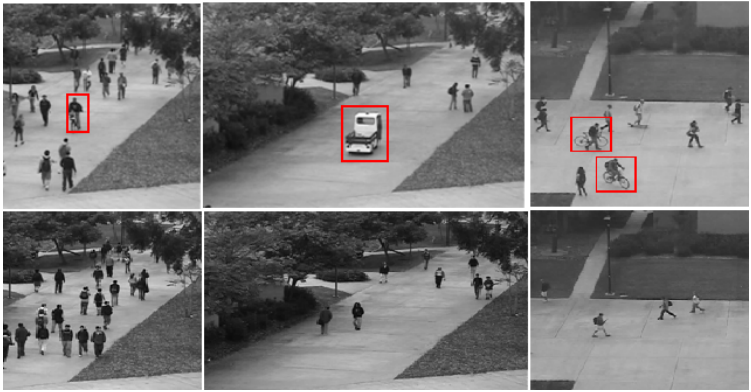

Figure 9: top and bottom row contain outliers (identified in a rectangular box) and non-outliers frames of UCSD anomaly data respectively.

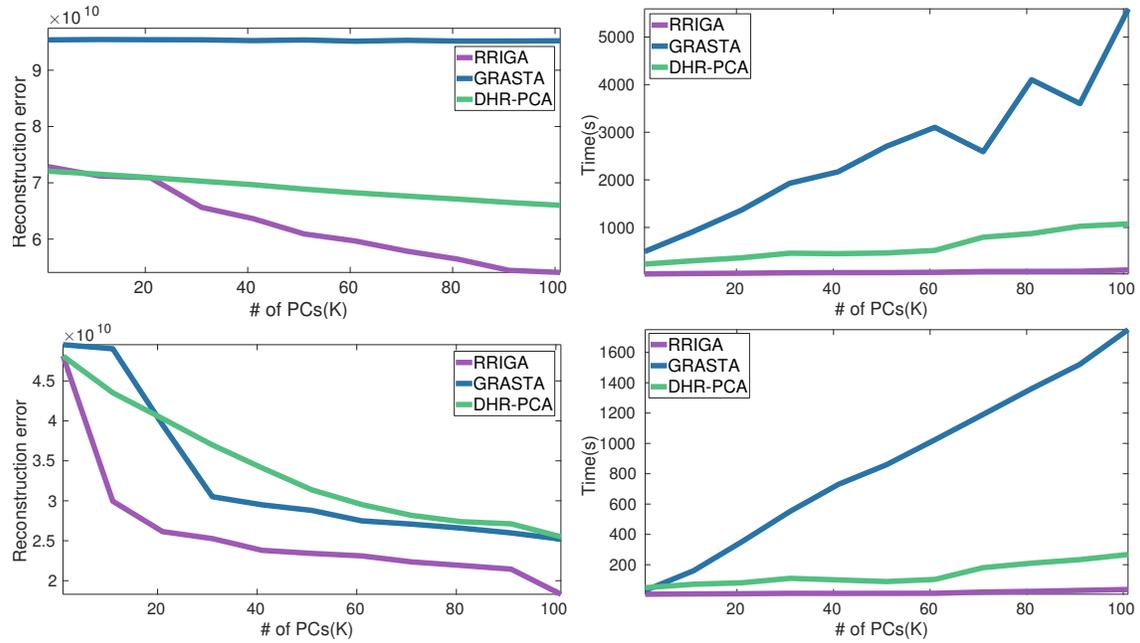

Figure 10: Performance of RRIGA, GRASTA and DHR-PCA on anomaly data; *Top two:* "Peds1"; *Bottom two:* "Peds2".

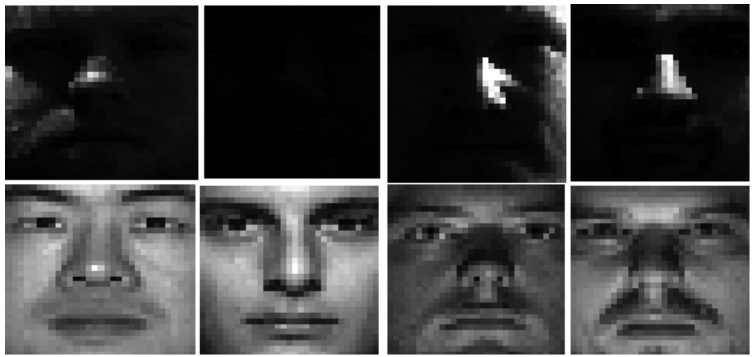

Figure 11: Top and bottom rows contain outliers and non-outliers images from YaleExtendedB data respectively.

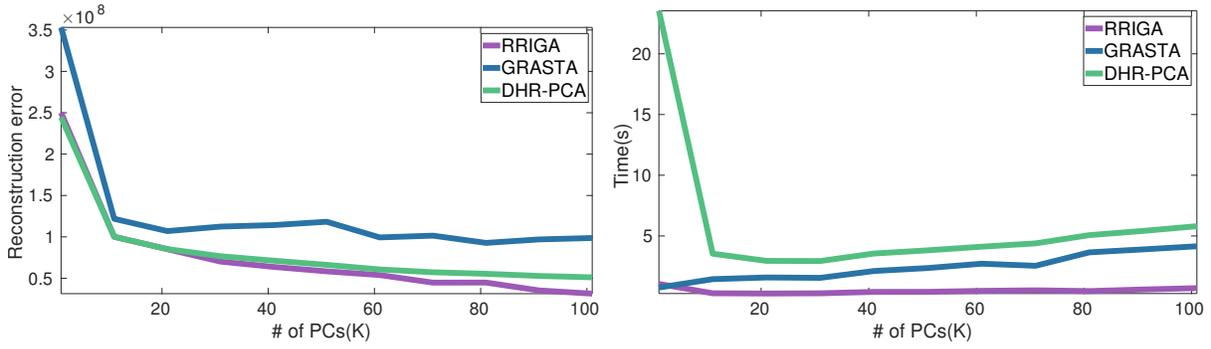

Figure 12: Performance comparison on YaleExtendedB data.

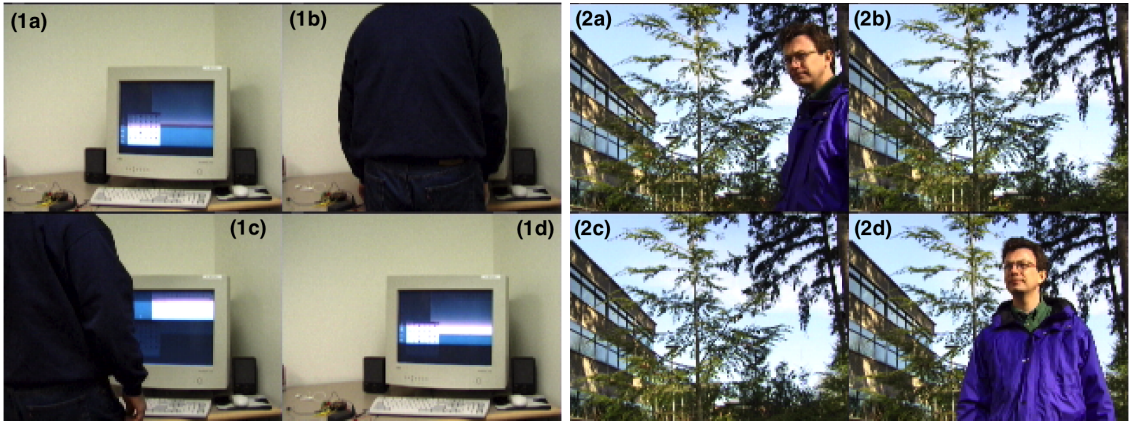

Figure 13: Image sets (1) and (2) contain training samples from the 'Camoflouge' and 'Waving Tree' data sets respectively. In these figures (1b), (1c), (2a), (2d) are the samples with outliers.

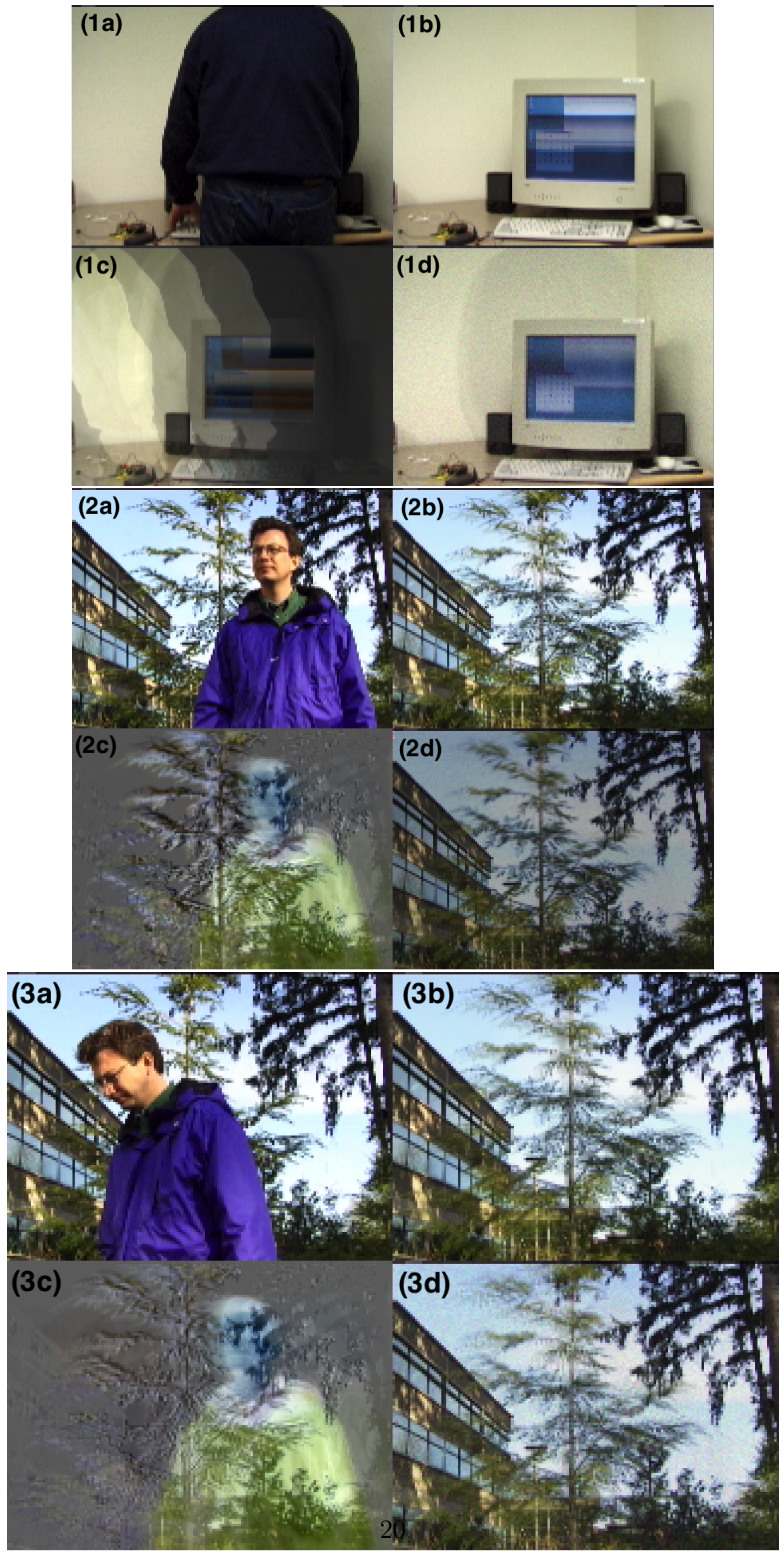

Figure 14: Image sets (1) and (2) contain 5% outliers and (3) has 10% outliers. The subfigure (a) is the test image and (b) RRIGA output (c) PCA output (d) DHR-PCA output.